\documentclass[11pt,a4paper,twocolumn]{article}
\usepackage{graphicx}
\usepackage[hidelinks]{hyperref} 
\usepackage{times}
\usepackage{fancyhdr}
\usepackage{xcolor}
\usepackage{amsmath}

\begin{document}
\voffset=0cm
\topmargin= 0cm
\headheight=0.5cm
\headsep=0.5cm
\hoffset=0cm
\marginparwidth=0cm
\marginparsep=0cm
\oddsidemargin=-0.4cm
\evensidemargin=\oddsidemargin
\textwidth=17.0cm 
\textheight=24.1cm 
\pagestyle{fancy}
\fancyhf{}
\fancyhead[EL,OL]{Journal of Virtual Reality and Broadcasting, Volume n(200n), no. n}
\fancyfoot[EL,OL]{urn:nbn:de:0009-6-348, ISSN 1860-2037}
\renewcommand{\bottomfraction}{0.5}
\renewcommand{\footrulewidth}{0.5pt}
\renewcommand{\headrulewidth}{0.5pt}

\title{Towards a Pipeline for Real-Time Visualization of Faces for VR-based Telepresence and Live Broadcasting Utilizing \\Neural Rendering}
\author{Philipp Ladwig\footnotemark[1] ,
Rene Ebertowski\footnotemark[1] , Alexander Pech\footnotemark[1] , Ralf Dörner\footnotemark[2] , Christian Geiger\footnotemark[1] \vspace{2mm} \\
\footnotemark[1] University of Applied Sciences Düsseldorf, Germany\\
Mixed Reality and Visualization Group (MIREVI)\\ \{\tt philipp.ladwig, rene.ebertowski,\\ \tt alexander.pech, geiger\}@hs-duesseldorf.de\\
\href{http://www.mirevi.de}{{\tt www.mirevi.de}}\vspace{2mm}\\
\footnotemark[2] RheinMain University of Applied Sciences\\
Faculty of Design – Computer Science – Media, Wiesbaden, Germany\\
 {\tt ralf.doerner@hs-rm.de}\\
\vspace{2mm}\\
}
\date{}

\maketitle
\thispagestyle{fancy}
\author{}
\date{}



\begin{table}[b]
\resizebox{\linewidth}{!}{
\begin{tabular}{|l|}
\hline
\textbf{Digital Peer Publishing Licence} \\
\hline
Any party may pass on this work by electronic\\
means and make it available for download under\\
the terms and conditions of the current version\\
of the Digital Peer Publishing Licence (DPPL).\\
The text of the licence may be accessed and\\
retrieved via internet at\\
\href{http://www.dipp.nrw.de/}{{\tt http://www.dipp.nrw.de/}}.\\
\hline
\end{tabular}
}
\resizebox{\linewidth}{!}{
\begin{tabular}{l}
\textit{First presented at the Workshop of GI Special Interest group VR/AR 2020,}\\ [-2pt]
\textit{extended and revised for JVRB}\\ [-2pt]
\end{tabular}
}
\label{jvrblicence} 
\end{table}

\begin{abstract}
While head-mounted displays (HMDs) for Virtual Reality (VR) have become widely available in the consumer market, they pose a considerable obstacle for realistic face-to-face conversation in VR since HMDs hide a significant portion of the participants faces. Even with image streams from cameras directly attached to an HMD, stitching together a convincing image of an entire face remains a challenging task because of extreme capture angles and strong lens distortions due to a wide field of view. Compared to the long line of research in VR, reconstruction of faces hidden beneath an HMD is a very recent topic of research. While the current state-of-the-art solutions demonstrate photo-realistic 3D reconstruction results, many of them require high-cost laboratory equipment and large computational costs. We present an approach that focuses on low-cost hardware and can be used on a commodity gaming computer with a single GPU. We leverage the benefits of an end-to-end pipeline by means of Generative Adversarial Networks (GAN). Our GAN produces a frontal-facing 2.5D point cloud based on a training dataset captured with an RGBD camera. In our approach, the training process is offline, while the reconstruction runs in real-time. Our results show adequate reconstruction quality within the ``learned'' expressions. Expressions not learned by the network produce artifacts and can trigger the Uncanny Valley effect.
\end{abstract}

\section{Introduction}
\begin{figure*} 
	\centering
	\includegraphics[width=\textwidth]{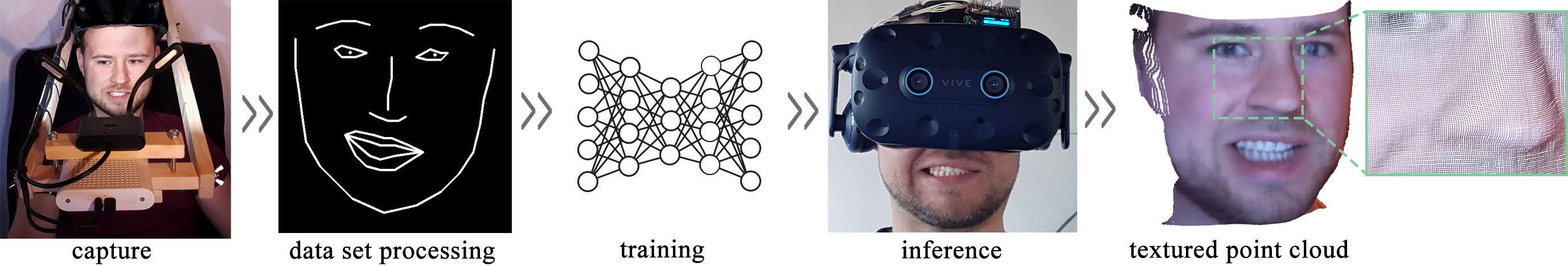}
    \caption{\textit{Our conceptual pipeline:
    First, we capture several RGBD images with a helmet camera mount. These images are processed and serve as the input data for our GAN. After training, the GAN produces textured point clouds in real time. In this work we improve the data set processing, training and inference stage compared to our previous systems\,\cite{Ladwig2020AufDemWeg, Ladwig2020Unmasking}. Building a face-tracking HMD is not part of the present work.}}
    \label{fig:banner}
\end{figure*}
\label{intro}

Natural face-to-face communication is three-dimensional. A conversation includes not only the verbal communication channel but also the non-verbal channel. In particular, eye contact, facial expressions as well as gestures performed with arms and hands (kinesics), and even the physical distance between each other (proxemics) are essential information carriers during a conversation\,\cite{Ladwig2018ALR}. Currently, common mainstream technologies for computer-mediated communication are video conferencing applications such as Skype or FaceTime. While these allow reading the facial expressions of the counterpart, no ``real'' eye contact is possible, wide gestures may be cut off in the camera image, deictic gestures are difficult to interpret spatially, and perceptual physical body distance between the participants does not exist.

Current head-mounted displays for VR are capable of delivering believable and immersive 3D experiences including telepresence. However, this does not fully apply to real-time social interactions in VR. When the face of a person is covered by an HMD, it is impossible to read its non-verbal facial communication cues, which is, in fact, a crucial communication channel between individuals. This is not only relevant for face-to-face meetings in VR (for example in VRChat\,\cite{vrchat}, Altspace\,\cite{altvr}, or Meta's Horizon Worlds\,\cite{metahorizonworlds}) but also in VR application scenarios in which only one VR user wears an HMD and tries to engage with their audience. For example, such a VR user could be an architect who presents ideas for a new building in VR to their clients, a Virtual YouTuber (VTuber), a Twitch streamer in front of a green screen who broadcasts themselves from inside a VR environment, or friends playing a VR game together in a living room.


The classic way for creating and rendering photo-realistic humans in real time is costly and requires a lot of manual effort such as scanning, modeling, and manual texturing from a skilled 3D artist. Furthermore, today's HMDs usually lack adequate sensors for face tracking. Only a few research groups have so far addressed this problem and presented methods that can generate authentic face avatars for VR without extensive manual modeling\,\cite{Thies18FaceVR, Lombardi18DAM, Wei2019VRFacial, Raj_2021_CVPR, Thies_2021_NeuralHeadAvatars}. These approaches are not available to the public, and many of them require expensive hardware\,\cite{Lombardi18DAM, Wei2019VRFacial}.

Human avatars and their perception have been studied in multiple domains. A systematic review of social presence concludes: "...multiple studies show that the vivid perceptions of another person often lead to greater enjoyment and social influence..."\,\cite{oh:2018:STARonSocialPresence}. Although, the Media Richness Theory\,\cite{mediaRichnessTheory} is almost 40 years old, recent studies still confirms it\,\cite{ishii2019revisiting}. It suggests that the most comprehensive exchange of information between people happens during face-to-face conversations compared to all other digital or analog communication possibilities. A higher quantity and quality of data shared typically leads to more effective communication. A key issue in this context of face-to-face telepresence in VR is the occurrence of the Uncanny Valley effect\,\cite{Mori70}. Humans are markedly sensitive to minimal and unnatural discrepancies in faces. As soon as a virtual human does not perfectly resemble a real human, it is often subconsciously classified as unlikeable, unpleasant, or even creepy. 

One technique that has successfully bridged the Uncanny Valley in recent years is Generative Adversarial Networks (GANs). Today, GANs serve as the core technology behind \textit{Deepfakes}. They enable such authentic results that their methods and algorithms are the subject of current research to distinguish fake images from real ones, as the human eye is no longer able to reliably do so\,\cite{roessler2019FaceForensic}. Therefore, we use algorithms in the context of this work that are also employed to create Deepfakes. We extend this approach with an additional dimension\,(textured 2.5D point cloud instead of only an RGB image) to generate realistic representations of 2.5D face avatars. We do not create a full 3D head because we only capture the face of a person from a static frontal position with an RGBD sensor. This implies that we do not generate realistic textures from side views. However, we maintain a stereoscopic perception of the reconstructed face during face-to-face conversations in a virtual environment.


We present an end-to-end learning system that has low hardware cost compared to others, requires moderate computational resources, and generates results with frame rates suitable for VR applications. Our research contributes to GANs playing a key role in authentic 3D telepresence applications in the near future. In addition to sharing our insights in this paper, we make the code of our prototype publicly available under: \url{https://github.com/Mirevi/face-synthesizer-JVRB}.

This work is an extension and improvement of our previous neural rendering pipeline\,\cite{Ladwig2020AufDemWeg} and complements our recent work on how to build an HMD with face tracking capabilities \cite{Ladwig2020Unmasking}. The contributions of this work are the creation of a face capture pipeline as well as the introduction of novel Generative Adversarial Networks (GAN)\,\cite{goodfellow2014generative} that are tailored for the authentic reconstruction of faces in three-dimensional telepresence and live broadcasting applications. The motivation is to use a commodity graphics card to capture and reconstruct the individual characteristics of a person's face with a high level of (personal) details and VR-enabled frame rates in order to create an authentic avatar that goes beyond the capabilities of today's avatar creation tools such as VRChat\,\cite{vrchat}, Altspace\,\cite{altvr} or Meta's Horizon Worlds\,\cite{metahorizonworlds}.




\section{Related Work}
Face reconstruction for telepresence and (live) broadcasting with an HMD occluding a person's face is a young research field. Olszewski et al.\,\cite{Olszewski16HighFid} presented a system that uses an RGB camera to transfer facial expressions from the lower face area to an avatar. Li et al.\,\cite{Li15FacialHMD} extended this approach with pressure sensors placed in the foam of an HMD capturing a person's facial expressions. The idea of using sensors within the HMD is similar to our concept, but we use personalized avatars that are trained in advance and synthesized in a final step.

Casa et al.\,\cite{Casas16Rapid}, Früh et al.\,\cite{Frueh2017}, and Thies et al.\,\cite{Thies18FaceVR} used stationary RGBD cameras to create personalized avatars of users. In the first step, the user was captured by the camera without an HMD in order to create a virtual avatar. When the user wore the HMD, the stationary RGBD camera recognized facial expressions. Due to the fixed position of the camera, the range of head motion was limited. Eye movements were registered by eye-tracking cameras and transferred to the user's face avatar. The approaches of Casa et al.\,\cite{Casas16Rapid} and Früh et al.\,\cite{Frueh2017} evoked the Uncanny Valley effect to varying degrees. To mitigate this, Früh et al.\,\cite{Frueh2017} did not completely remove the HMD, but rendered it as a semi-transparent object. These systems are similar to our approach in the way that they create a personalized avatar using an RGBD camera and produce almost photo-realistic avatars. The approach of Thies et al.\,\cite{Thies18FaceVR} demonstrated better results by using a 3D morphable model (3DMM) as underlying head mesh template and as an inductive bias to their system that provides fundamental data about the composition of a human face. They optimize the visual quality by an \textit{analysis-by-synthesis} approach\,\cite{3dmm} and achieve photo realistic results with only a few image artifacts. While the visual quality is convincing, this approach only provides stereoscopic renderings without the ability to freely choose the perspective around the reconstructed face because the final results are based on a given 2D video. Furthermore, it inherently does not allow for manipulation of the head's rotation, scale, and position in the final result.

The systems of Lombardi et al.\,\cite{Lombardi18DAM}, Wei et al.\,\cite{Wei2019VRFacial}, and Raj et al.\,\cite{Raj_2021_CVPR} create photo-realistic avatars with authentic facial expressions. While previous works completed the generation of personalized avatars within a few minutes, the system of Lombardi et al.\,requires computational time of more than a day. The three-dimensional avatar is generated with the aid of a large number of high-resolution images from different angles and facial expressions with an expensive hardware setup that generates a large amount of data for further processing. The created face avatar can be controlled by three RGB cameras attached to an HMD. A key component of this system is the use of \textit{Variational Autoencoders} (VAEs). Both VAEs and GANs have been proven several times to be suitable for authentic face reconstruction. However, since literature shows that VAEs and only a L1 loss tend to produce blurry results more often, we use GANs\,\cite{HandsOnGANs}. The latter concept was first presented by Goodfellow et al.\,\cite{goodfellow2014generative}, and Radford et al.\,\cite{radford2015unsupervised} improved it in a sustainable way. Furthermore, Karras et al.\,\cite{karras2017progressive} achieved photo-realistic portrait images that are indistinguishable from real photographs by using the principle of \textit{Progressive Growing GAN}. However, according to Karras et al., the GAN has little to no external control over the appearance of the generated object or face because the input to the network is a latent vector without any direct relation to a face property such as hair color, facial expression, or gender. In further works Karras et al.\,\cite{karras2018stylebased} enhanced the architecture of the GAN and were able to automatically separate higher-level attributes (e.g. pose, identity) from stochastic variations (e.g. freckles, hair). Nevertheless, this approach does not allow to explicitly control the facial expression.

Conditional GANs (cGANs) have been shown to be able to learn and reproduce specific relationships between inputs and outputs that are understandable for humans. For example, Mirza and Osindero\,\cite{mirza2014conditional} have extended the input to the generator and discriminator with a label ${y}$, which makes it possible to generate images from a particular category ${y}$. This method for conditioning GANs was developed further by Radfort et al.\,\cite{radford2015unsupervised} with the DCGAN and by Isola et al.\,\cite{pix2pix2016} with the Pix2Pix GAN. They replaced the noise input vector $z$ with a user-defined input vector. Without a noise vector, there is no latent space $Z$ (since ${z \in Z}$). If the stochastic aspect contained in the noise vector is not compensated, the GAN will only memorize the training examples. Any inputs that deviate from the training data would lead to inadequate results, as described by Isola et al.\,\cite{pix2pix2016}. By using a U-net architecture\,\cite{u-net} with dropouts in the Pix2Pix GAN, the stochastic aspect as well as the missing latent space can be otherwise integrated into the generator. The discriminator of the Pix2Pix GAN receives the same input image $x$ as the generator as well as its output image $y_{fake}=G(x)$ or the image $y_{real}$ matching $x$ from the dataset. This is basically equivalent to the idea of cGANs\,\cite{mirza2014conditional} where not only the output of the generator is evaluated but also its difference from the input. Unlike the cGAN, the output of the discriminator of the Pix2Pix GAN is not a scalar that decides between ``real''  or ``false'' but a matrix. By using convolutional layers (cf. Radford et al.\,\cite{radford2015unsupervised}), each entry in the output matrix represents an $n*m$-sized region (so-called patch) of the input image. This allows abstract representations to be admitted as matrices (e.g. images) for conditioning the network to have a controlled influence on the output of the generator. This approach was further developed by \cite{wang2017highresolution} with the Pix2PixHD GAN to generate images with a higher resolution and more details. In this paper, we adapt the idea of cGANs, especially of the Pix2Pix and Pix2PixHD frameworks, and tailor them to our application domain.
\section{System}
In the following, we explain the process and structure of the proposed system, as shown in Fig.\,\ref{fig:banner}, and then discuss the steps in more detail in the subsequent sections. 

Our process starts with the acquisition of a personal RGBD dataset. The acquired data is preprocessed by an automated procedure. A Facial Landmark Map (FLM) per RGB image is extracted and saved beside the corresponding RGB image. It decodes the facial expression of the respective RGB image in a binary image as so-called landmarks as shown in the second image from left in Fig.\ref{fig:banner}. Our proposed GAN is trained with the captured RGBD images as well as with the corresponding FLM. For each person our GAN must be trained from scratch. We do not use any inductive biases like a 3DMM\,\cite{3dmm} and the system does not learn correspondences between persons. After the training, the system can be used for real-time telepresence or live broadcasting. In the VR application scenario, the user would wear a face-tracking head-mounted display that could create an FLM in real time, which we then feed into the trained generator module of our GAN. This paper does not focus on building and implementing a face-tracking HMD. Further hardware-related implementation details are described in our previous work\,\cite{Ladwig2020Unmasking}. The GAN could create an RGB and a D image of the ``learned'' person based on the FLM, and finally, we fuse the generated RGB and D images into a textured point cloud.

\subsection{Training Data}
\begin{figure}[htbp]
\begin{center}
	\includegraphics[width=0.5\textwidth]{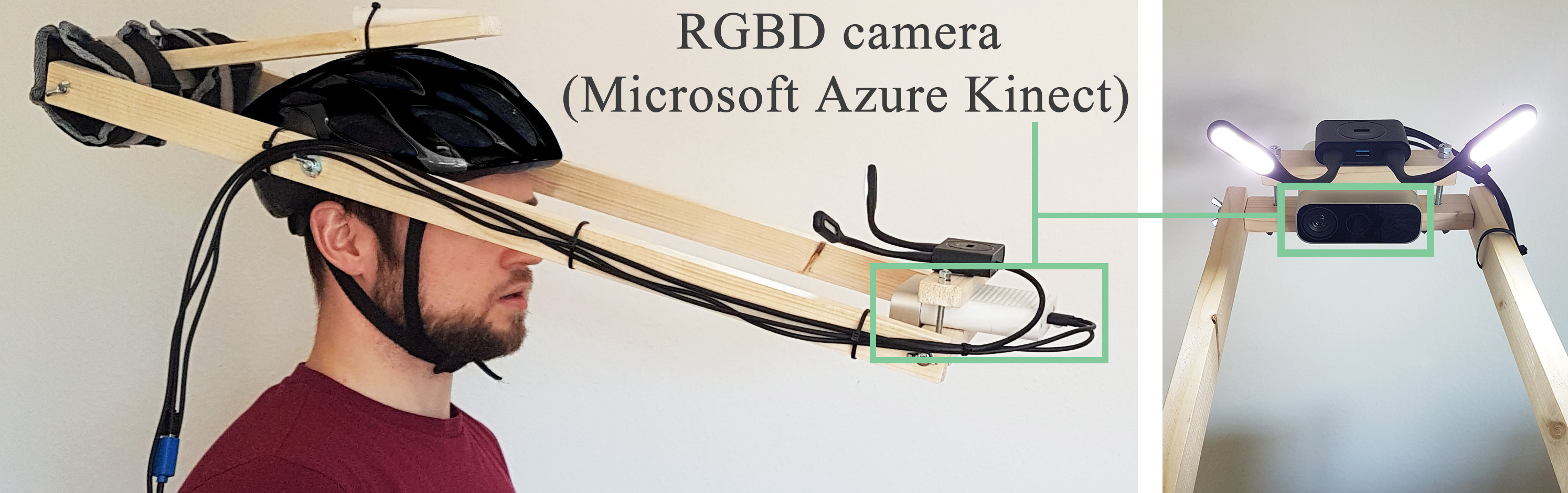}
    \caption{\textit{Helmet camera mount for RGBD data acquisition. This mount ensures that head rotations are not included in the training dataset and, therefore, reduces the entropy in the dataset. Moreover, this method significantly reduces the training time and increases the visual quality of the output images. The material price of the helmet mount without RGBD camera is about 60\,USD.}}
    \label{fig:helmet}
    \end{center}
\end{figure}
An RGBD dataset of the respective person forms the basis for the training process. For the acquisition process, an RGBD camera is mounted in a fixed position to a self-made helmet mount, as shown in Figure \ref{fig:banner} left and Figure \ref{fig:helmet}. This mount ensures that the entropy in the dataset is as minimal as possible. Varying distances, positions and head rotations do not contribute to ``learning'' the user's face. By using the helmet mount, we ensure that the training time of the networks is short and the reconstruction quality of the final results is high, as we learned from previous experiments conducted without the helmet mount. 

The RGBD sensor in the mount captures facial expressions and stores the color information as a 3-channel image file (8-bit for each channel) with 2048 $\times$ 1536\,pixels, while it stores the depth information as a 1-channel image file (16-bit grayscale) with 640 $\times$ 576\,pixels. During the capture process, the user is encouraged to perform a variety of different facial expressions. A dataset of a person should contain about 1500 to 2000 RGBD images, and the acquisition process takes about 10\,minutes. The data is then preprocessed for further steps. First, the foreground and background pixels are clipped in front of and behind the face and the depth resolution is reduced from 16 to 8 bit, which reduces the depth range from 65,535 to 255\,mm. This speeds up the training process of the GAN and significantly reduces depth noise. With the help of the helmet mount, we ensure that a depth range of 255\,mm is sufficient for spatially reproducing the frontal part of the head since the distance between the sensor and the face in the helmet mount is always fixed. Furthermore, the data is normalized and the image data is converted into values between -1 and 1 for the training process. Contrasts are sharpened by normalizing the histogram.  

\subsection{Determination of Facial Landmarks}
To control the output of the generator and thus the facial expressions of a person's face, the dataset must be labeled before the training process. We use 70 facial landmarks (68 of the Multi-PIE scheme\,\cite{multiPIE2010} and two for the location of the irises) as binary images for each tuple of RGB and corresponding depth image in the dataset. As mentioned before, we call these binary images Facial Landmark Maps (FLM). The position of the landmarks identifies the expressions of the person in each image of the dataset. To determine the landmarks, we use the \emph{Face Alignment Network} (FAN) of Bulat and Tzimiropoulos\,\cite{Bulat_2017}. The FAN is not able to determine the position of the person's pupils within an image. Therefore, two additional landmarks were implemented based on an eye-tracking procedure by Timm and Barth  \cite{Timm2011AccurateEC, pupilDetector}. We experimented with an Tobii Eye Tracker 4C, but users reported that additional weight on the helmet mount felt uncomfortable during the capture procedure. Both the tracking results of the Tobii Eye Tracker and Timm and Barth were similar and sufficient for our application.

When the landmarks in each image had been located, a rectangle of the maximal and minimal locations of the landmarks was created and the RGB and D images were cropped to this area and resized to $512 \times 512$\,pixels.

\subsection{Neural Network Architecture}
Previous work with neural networks, such as Wu et al.\,\cite{3dgan}, has shown that a voxel-based approach is associated with high training and execution times of the model. Therefore, an RGBD-based solution was targeted. The advantage of this approach lies in the compact representation of the data as a point cloud and the possibility to adapt previous RGB-based methods. Our underlying network architecture is derived from the Pix2Pix GAN by Isola et al.\,\cite{pix2pix2016}. In earlier experiments, we discovered that this architecture was able to produce acceptable results\,\cite{Ladwig2020AufDemWeg, Ladwig2020Unmasking}, but the images have a low resolution of $256 \times 256$\,pixels, often lack details in high frequency areas such as facial hair and tend to produce time-inconsistent reconstructions with noise. Therefore, we experimented with the Pix2PixHD framework\,\cite{wang2017highresolution}, which is an extension of the Pix2Pix GAN\,\cite{pix2pix2016}. Although it produces images with a higher resolution and better quality, the inference time is not capable of retaining real-time frame rates on commodity hardware. Therefore, we kept the Pix2Pix framework as a base and gradually added elements from the Pix2PixHD framework and further improvements from other works until we obtained an acceptable image quality with a reasonable processing speed for interactive frame rates. In summary, we propose the following changes to the Pix2Pix framework:
\begin{enumerate}
\item We added a multi-scale discriminators that receives three different resolutions of the input image and an additional \emph{Feature Matching Loss} as described in Pix2PixHD\,\cite{wang2017highresolution}.
\item We changed the \emph{Sigmoid Cross Entropy Loss} of Pix2Pix's discriminator to the \emph{Least-Squares Loss} of the LSGAN\,\cite{Mao2017LSGAN} as suggested by Wang et al.\,\cite{wang2017highresolution}.
\item We exchanged the \emph{Perceptual-VGG Loss}\,\cite{PercLossJohnson2016} originally suggested by Wang et al.\,\cite{wang2017highresolution} with the better performing \emph{Learned Perceptual Image Patch Similarity} (LPIPS) by Zhang et al.\,\cite{zhang2018perceptualLPIPS}.
\end{enumerate}

For the discriminator we obtain the following objective:
\begin{equation}\label{eq:lsgan_objective1}
		\min_{D_1, D_2, D_3} V_{GAN}(D) = \sum_{k=1, 2, 3}\mathcal{L}_{cLSGAN\_D}(D_k, G) 
\end{equation}
where $D_1$, $D_2$, and $D_3$ describe the three different resolution of the input image. For the generator we end up with the following objective function:
\begin{equation}\label{eq:lsgan_objective2}
\begin{split}
        & \min_G V_{GAN}(G) = \\
        & \sum_{k=1, 2, 3}\Bigl[\mathcal{L}_{cLSGAN\_G}(D_k, G) + \lambda_{FM}\mathcal{L}_{FM}(D_k, G)\Bigr] \\
        & + \lambda_{L1}\mathcal{L}_{L1}(G) + \lambda_{LPIPS}\mathcal{L}_{LPIPS}(y, G(x))
\end{split}
\end{equation}
where we choose the following hyper-parameters:
$\lambda_{FM} = 10$, $\lambda_{L1} = 100$, $\lambda_{LPIPS} = 10$. The functions $\mathcal{L}_{cLSGAN\_D}$ and $\mathcal{L}_{cLSGAN\_G}$ can be found in Mao et al.\,\cite{Mao2017LSGAN}, whereas the function $\mathcal{L}_{FM}$ is based on the feature matching loss of Wang et al.\,\cite{wang2017highresolution}. In order to maintain faster inference time than the Pix2PixHD we did not implement the coarse to fine approach for the generator. We sacrifice very high resolution of the output images for computational speed. We provide further details about the architecture on GitHub.

Using these improvements helps to prevent high-frequency details such as facial hair and significantly increases the reconstruction quality, as explained further in the results section (sec.\,\ref{sec:results}). Because we mainly enhanced the loss function and the discriminator side, we did not need to change the generator. Therefore, we are able to maintain high frame rates during inference since only the generator module is used in the telepresence and live broadcasting scenario. As a side effect, the training process requires more memory, but the overall training time decreases by more than a half compared to our Pix2Pix-only approach (from about 19\,hours to 8\,hours) because the new loss term is more purposeful for our application and, therefore, helps to obtain better results in less time.

\begin{figure}
	\centering
	\includegraphics[width=0.5\textwidth]{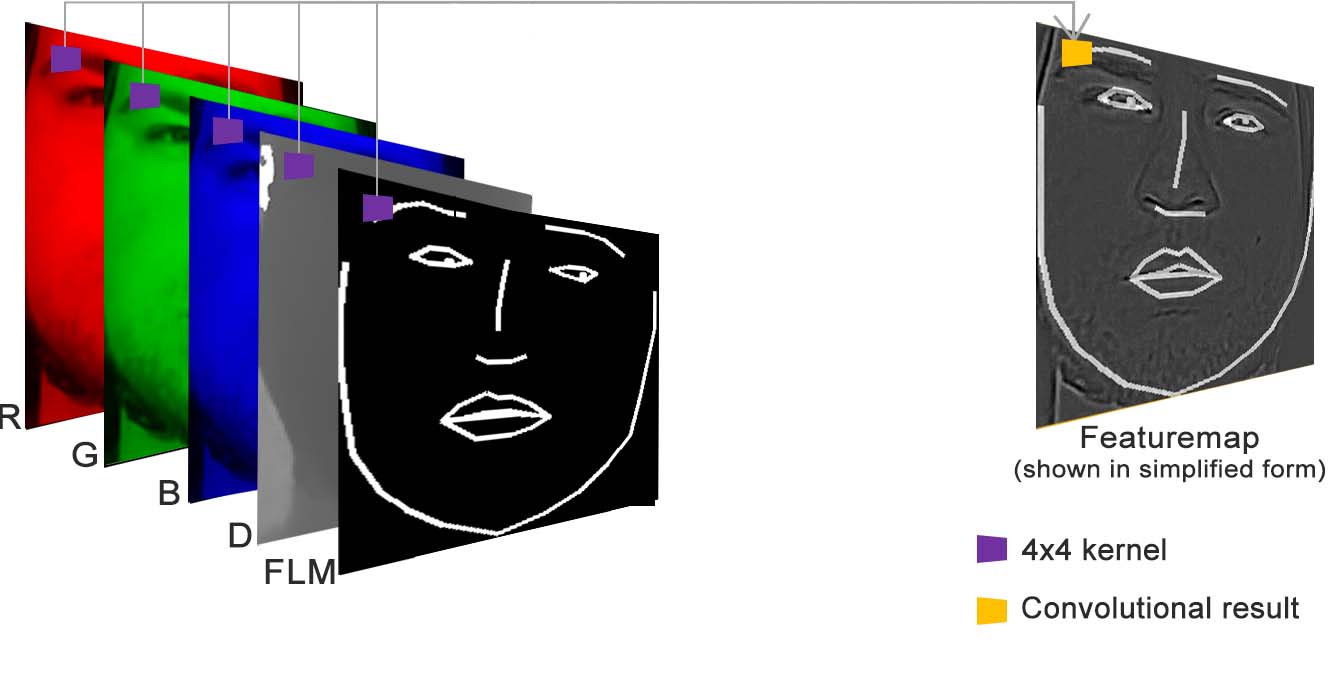}
    \caption{\textit{Example convolution for the discriminator input. Each RGBD channel and the FLM are weighted individually.}}
    \label{fig:InputDis}
\end{figure}

The generator of our GAN receives a $512 \times 512$\,pixel FLM of the facial landmarks as input. Compared to the Pix2Pix GAN, the output has been extended by a fourth feature map to be able to generate depth images. In addition, the discriminator receives five feature maps as input instead of only four. While the first four correspond to the four channels of the RGBD image, the remaining feature map contains the corresponding FLM, as visualized in Fig. \ref{fig:InputDis}. One of our early hypotheses in \cite{Ladwig2020AufDemWeg} was that a higher number of FLMs (e.g.\,four times) would result in better reconstruction results to start the training process with a balanced ratio between RGBD images and FLMs (cf.\,Fig.\,\ref{fig:InputDis}). This hypothesis has been disproved. Changing the number of FLMs does not change the quality of the results but only increases the training time.
\subsection{Training Process}\label{sec:trainingProcess}
For each person, the discriminator and the generator are each completely trained from scratch. Before the training process, all weights of the two networks are initialized with a random value. The random values follow a Gaussian distribution with an expected value of $0$ and a standard deviation of $0.02$. The batch size of the input data into the GAN is $1$, and the epoch count is $100$. Both generator and discriminator are trained with an initial learning rate of $0.0002$, which decreases linearly towards zero over the last 70 epochs. The discriminator $Loss_D = (Loss_{Dreal}+Loss_{Dfake})*0.5$ reduces the learning rate, making it slower to learn relative to the generator. This is necessary because at the beginning of the training phase the discriminator can effortlessly accomplish its task. If the discriminator learns too quickly, the generator has no chance to learn how to create the desired face.


\section{Results}\label{sec:results}
\begin{figure*} 
	\centering
	\includegraphics[width=\textwidth]{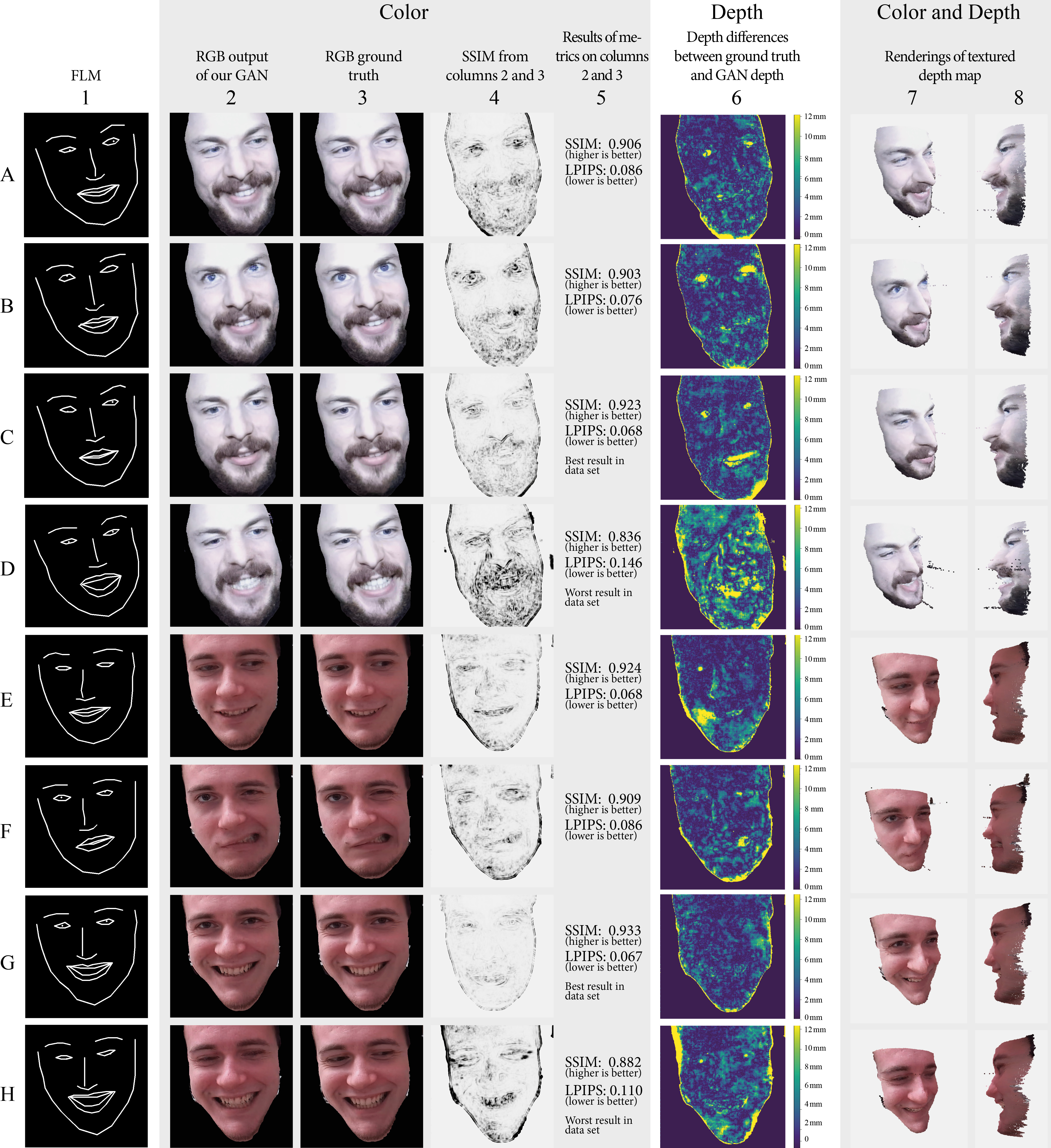}
    \caption{\textit{Results 1/2: This overview shows FLMs in column 1 from our evaluation datasets, so the results are based on unseen data for the neural network. The FLMs were created from the images in column 3 -- a real image from the evaluation data set. Column 2 shows the results generated by our GAN. The GAN received the FLM from column 1 and generated the images in column 2. Column 4 depicts the SSIM difference. Darker values indicate larger differences between the images in columns 2 and 3.  Column 6 visualizes the error between the generated depth and the ground truth depth. The combination of the generated depth and color data can be seen in columns 7 and 8 from an angle of 30 and 90\,degrees.}}
    \label{fig:results}
\end{figure*}

GANs are difficult to train because of their adversarial training procedure to find the balance between the learning rate and losses of the generator and discriminator networks. As we use many different losses, a long line of hyper parameter tuning was necessary to find the optimal settings. Fig.\,\ref{fig:results} shows the reconstruction results for four different persons with the best training parameters described in section \ref{sec:trainingProcess}. 

We did not use the face-tracking HMD for the evaluation, similar to \cite{Ladwig2020Unmasking}, because it can cause slight tracking errors and could decrease the quality of the results for comparison. Our intention is to directly compare the improvements of our pipeline to our previous system described in \cite{Ladwig2020AufDemWeg}. For a comparison in motion, we refer the reader to the corresponding video that can be found via the GitHub link.

\begin{figure}
	\centering
	\includegraphics[width=0.5\textwidth]{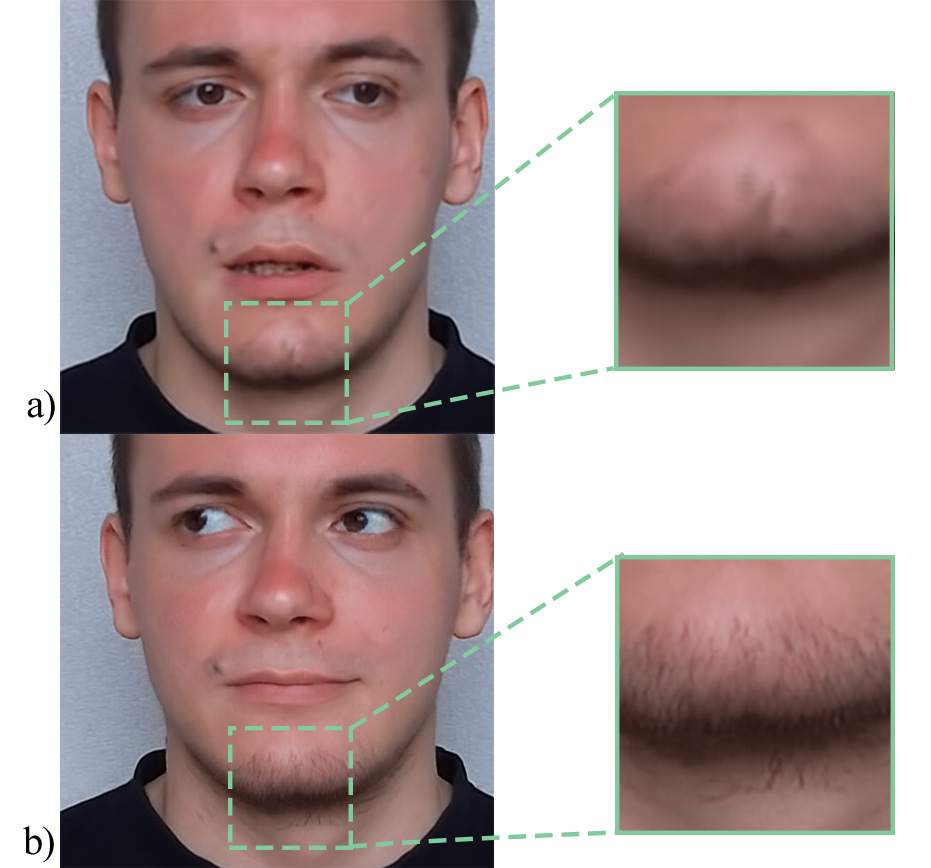}
    \caption{\textit{Our new pipeline, network architecture and losses significantly improved the quality. Image a) shows a sample from the previous system of Ladwig et al.\,\cite{Ladwig2020AufDemWeg, Ladwig2020Unmasking}. Image b) illustrates the enhanced resolution (from $256 \times 256$ to $512 \times 512$\,pixels) and the improved preservation of high-frequency details.}}
    \label{fig:sharperResults}
\end{figure}
As a quantitative metric for the assessment of the reconstruction quality, we use \emph{Structural Similarity} (SSIM)\,\cite{ssim} and \emph{Learned Perceptual Image Patch Similarity} (LPIPS)\,\cite{zhang2018perceptualLPIPS}. Our previous system\,\cite{Ladwig2020AufDemWeg} performed on average with an SSIM of 0.851 and a value of 0.114 on LPIPS. Our proposed system reaches on average 0.910 for SSIM (higher is better) and 0.082 for LPIPS (lower is better). The numerical results of the SSIM metric are comparable to a JPEG compression of about half the original file size of the images in column 3. In contrast, the previous system\,\cite{Ladwig2020AufDemWeg} only achieved reconstruction quality of less than a quarter of the original file size by means of a comparison with the JPEG compression.

\begin{figure*} 
	\centering
	\includegraphics[width=\textwidth]{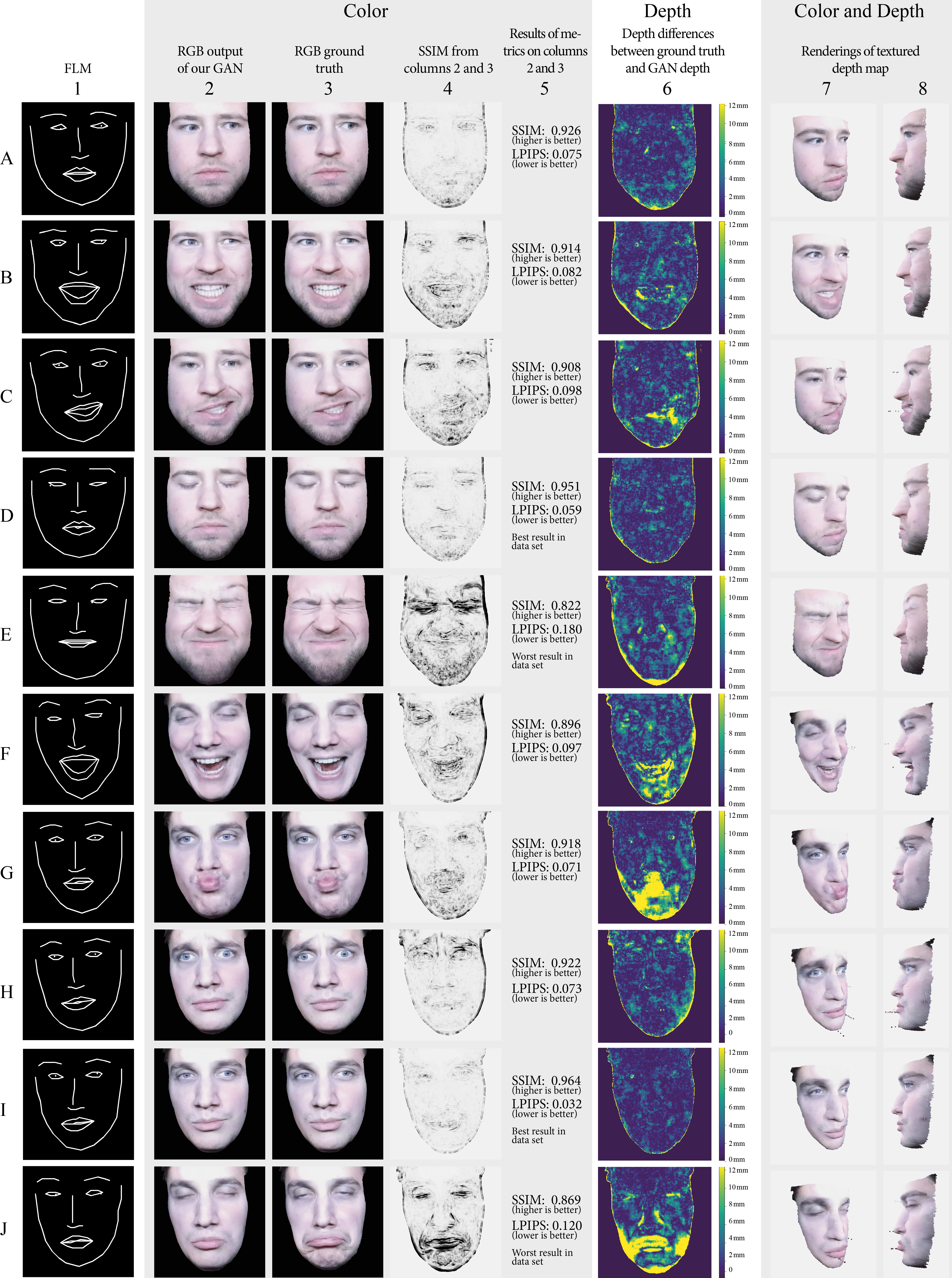}
    \caption{\textit{Results 2/2: Two further subjects are shown. The two last samples of each person summarizes the best and the worst images (measured by SSIM and LPIPS) and reflects the range of the reconstruction quality.}}
    \label{fig:results2}
\end{figure*}

All results and metrics shown in Fig.\,\ref{fig:results} and \ref{fig:results2} are generated from LFMs from the test set. This means, no backpropagation was conducted on the test set. The train/test split was 85\% to 15\%. The dataset for the first participant in Fig.\,\ref{fig:results} and \ref{fig:results2} was split 1238/207, for the second 1500/250, for the third 1620/271 and for last 2413/403. Please note that the dataset for the last participant was around 1000 items larger compared to the previous participants and lead to slightly better quantitative results for the best and worst result according to SSIM and LPIPS (the last two images per participant). This allows the conclusion that a larger data set tends to lead to better image quality in our scenario.

The main issue of our previous approach was the sharpness of the generated images\,\cite{Ladwig2020AufDemWeg, Ladwig2020Unmasking}, as shown in Fig.\,\ref{fig:sharperResults}. Due to the new architecture and loss functions, the system produces images with more details. Even skin pores are reconstructed well, e.g. on the user's forehead, as can be seen in Fig.\,\ref{fig:results}, rows E to H. Furthermore, we noticed a better reconstruction quality in areas with high-frequency details such as facial hair, as shown in Fig.\,\ref{fig:sharperResults}. Also, the temporal consistency is improved. Please see the linked video on our GitHub page for details.

The error between the generated and ground truth depth values is mostly below 4\,mm, as depicted in column 6. Exceptions are the reconstruction results with the worst SSIM and LPIPS metrics per dataset of a person, such as shown in Fig.\,\ref{fig:results}, row D and H, as well as in Fig.\,\ref{fig:results2}, row E and J. Furthermore, outliers can be seen in column 8. Note that we use the raw depth image of the Kinect and the raw output of our GAN. We do not filter or smooth the images, therefore, we assume that the network has also learned the depth noise of the sensor, which causes additional depth errors.

To compare the faces between the images of columns 2 and 3 without measuring background changes, we determined the facial area based on depth values and rejected all other pixels. At the border area between the faces and the background, large differences in the SSIM and the depth difference visualization can be seen. These differences are caused by the fact that the cropped areas of the real and generated images do not always align perfectly due to minimal differences in the generated faces. In addition, we also apply erosion and clipping to the face to reject parts of the background, which can cause the minimal differences.

Although our quantitative metrics indicate better results, our system still shows limitations in the reconstruction quality of the eyes, lips and oral cavity. Especially teeth and the tongue are often reconstructed with noisy artifacts, as can be seen in Fig.\,\ref{fig:results}, row D, column 2. The error increases with a graceful degradation when the expression moves towards exaggerated expressions that are far from the neutral face expression. The eyes are reconstructed with less artifacts than the oral cavity, but we observed that even a little amount of image artifacts can evoke the Uncanny Valley Effect, as can be seen especially in Fig.\,\ref{fig:results}, row B, column 2. Please zoom in for details.


In our experiments we used PyTorch 1.8 and Python 3.7 on Windows 10. We trained the same data set with approximately 1600 elements (an element is a set comprised of an RGBD image and an FLM) on two different machines. This took 8\,hours on a system equipped with an Nvidia RTX2080Ti, an AMD Ryzen Threadripper 2990WX and 128GB RAM. With a newer system, comprised of an Nvidia RTX3090, an Intel i7-9900K and 32GB RAM, training took only 4\,hours and 13\,minutes. Our previous system took 17-20\,hours on the machine with the RTX2080Ti for only 600 elements. 

The time required for a forward pass of the generator module with an image size of $512 \times 512$\,pixels is between 3 and 4\,ms (333--250\,fps) on an Nvidia RTX3090 and between 6 and 7\,ms (167--143\,fps) on an Nvidia RTX 2080. Our previous system was faster (between 1 and 3\,ms) but generates only images with a size of $256 \times 256$\,pixels. The timings of the present system are still suitable for VR-based application, where 75 to 120\,fps are common frame rates. However, note that many face tracking systems only work with 60 or even 30\,fps, which can limit the frame rate of the pipeline.

\section{Limitations and Future Work}
As mentioned before, one of the major issues is the low reconstruction quality of exaggerated expressions of the eyes, lips and oral cavity. The artifacts can induce the Uncanny Valley Effect and must be avoided for telepresence or broadcasting applications. As a further step, we plan to use a 3DMM\,\cite{3dmm} such as the FLAME model\,\cite{Li:FLAME:SiggraphAsia2017} as a better inductive bias to regularize depth and color information more efficiently. Furthermore, we observed that landmark tracking is not sufficient for faithful lip movement during speech. Therefore, an additional input signal besides the landmarks is necessary. A conditioning of speech as audio signals could provide a solution. Another issue to improve is the uncomfortable helmet mount. A solution with a stationary RGBD camera placed on a tripod or table is a favorable approach for future research.
Although, we achieve real-time frame rates on a gaming GPU, the generator network could still be too slow for using on a current stand alone headset like a Meta Quest 2 or a Vive XR Elite. Further experiments, hardware and performance improvements could solve this problem in the future.


\section{Conclusion}
We presented an improved end-to-end pipeline compared to previous approaches\,\cite{Ladwig2020AufDemWeg, Ladwig2020Unmasking} and a new GAN architecture that can learn facial identity and individual expressions of a user and reproduce them as a textured point cloud with frame rates that are suitable for Virtual Reality, telepresence and broadcasting environments. We have incorporated and extended the architecture, losses, and processing pipelines of several approaches from the field of neural rendering. Compared to previous works, our proposed system generates higher quality image results with slightly longer run time at inference. We achieved this goal by mainly changing the architecture of the discriminator while keeping the architecture of the generator lean. The reconstruction results partially lie in the Uncanny Valley, but they still convince with an authentic visualization of the respective person's identity and individual facial expressions. We believe that neural rendering will be a crucial part of photo-realistic rendering of humans in real-time applications in the future. Our work is a further step into this direction and hopefully helps to understand, improve and apply this technology.

\section{Acknowledgments}
\label{ackno}
We thank the MIREVI group at the University of Applied Sciences Düsseldorf and the \textit{Promotionszentrum Angewandte Informatik (PZAI)} in Hessen, Germany. This work is sponsored by the German Federal Ministry of Education and Research (BMBF) under the project numbers 16SV8182 (HIVE-Lab), 13FH022IX6 (iKPT\,4.0) and 16SV8756 (AniBot).

\bibliographystyle{amsalpha-ext}
\bibliography{main}

\end{document}